\documentclass[runningheads]{llncs}

\usepackage[T1]{fontenc}
\usepackage{graphicx}
\usepackage{hyperref}
\usepackage{booktabs}
\usepackage{tabularx}
\usepackage{amsmath}

\begin{document}

\title{Inferring Chronic Treatment Onset from ePrescription Data: A Renewal Process Approach}

\author{
  Pavlin G. Poli\v{c}ar\inst{1} \and
  Dalibor Stanimirovi\'{c}\inst{2} \and
  Bla\v{z} Zupan\inst{1}
}

\titlerunning{Inferring Chronic Treatment Onset from ePrescription Data}
\authorrunning{Poli\v{c}ar, Stanimirovi\'{c}, Zupan}

\institute{
  Faculty of Computer and Information Science, University of Ljubljana, SI-1000 Ljubljana, Slovenia\\
  \email{pavlin.policar@fri.uni-lj.si}
  \vspace{2mm}
  \and
  Faculty of Public Administration, University of Ljubljana, SI-1000 Ljubljana, Slovenia
}

\maketitle

\begin{abstract}
Longitudinal electronic health record (EHR) data are often left-censored, making diagnosis records incomplete and unreliable for determining disease onset. In contrast, outpatient prescriptions form renewal-based trajectories that provide a continuous signal of disease management. We propose a probabilistic framework to infer chronic treatment onset by modeling prescription dynamics as a renewal process and detecting transitions from sporadic to sustained therapy via change-point detection between a baseline Poisson (sporadic prescribing) regime and a regime-specific Weibull (sustained therapy) renewal model. Using a nationwide ePrescription dataset of 2.4 million individuals, we show that the approach yields more temporally plausible onset estimates than naive rule-based triggering, substantially reducing implausible early detections under strong left censoring. Detection performance varies across diseases and is strongly associated with prescription density, highlighting both the strengths and limits of treatment-based onset inference.

\end{abstract}

\section{Introduction}

Longitudinal electronic health record (EHR) data are frequently left-censored due to the gradual introduction of digital healthcare systems and incomplete migration of historical records~\cite{AlSahab2024,Gurupur2025}. In Slovenia, national e-prescribing achieved near-complete coverage immediately after its introduction in 2016, whereas diagnosis recording was adopted more gradually, increasing from 350,000 entries in 2016 to over 10 million by 2020. These low early counts primarily reflect incomplete system adoption rather than true disease incidence. Because diagnoses are often recorded only once and not systematically updated, historical disease onset is frequently missing or unreliable. Diagnosis data may also contain inconsistencies, labeling and coding errors, with reported misclassification rates up to 20\%~\cite{Goldstein2025}. For example, double diabetes --- the co-occurrence of type 1 and type 2 diabetes --- is a documented clinical phenomenon. However, in our nationwide dataset of 2.4 million individuals, the co-occurrence rate is approximately three times higher than rates reported in the literature~\cite{Merger2016}. In contrast, prescription data are renewal-based and must be periodically re-issued, providing a continuous and more reliable signal of ongoing disease management that can complement incomplete diagnosis histories.

Treatment patterns provide a complementary signal for inferring disease onset, yet many phenotyping approaches rely on simple heuristics that ignore prescribing dynamics. In machine learning tasks, this can affect cohort construction and introduce data leakage, for example, when treatment signals precede recorded diagnoses~\cite{Chiavegatto2021,Starcke2025,Ramadan2025}. Robust onset inference is therefore important for reliable label construction and downstream AI/ML modeling on EHR data. Modeling prescription events as stochastic temporal processes allows us to detect chronic treatment onset as a shift in prescribing dynamics. Chronic noncommunicable diseases are particularly suitable for this approach: they account for up to 80\% of deaths in Slovenia~\cite{StaroveskiLesnik2025} and are characterized by sustained, structured prescribing patterns that provide strong longitudinal signals, whereas acute conditions typically generate sparse or short-lived treatment patterns that are harder to detect.

In this work, we infer treated-phenotype onset by modeling longitudinal prescription trajectories as a renewal process. We apply change-point detection to identify transitions from sporadic to sustained therapy, formulated as a transition from a baseline Poisson (sporadic prescribing) regime to a renewal-based Weibull (sustained therapy) regime, distinguishing true onset from administrative noise. We evaluate this approach against rule-based triggering, and show that renewal-based modeling produces more temporally plausible onset estimates while revealing disease-specific detection limits driven by prescribing density.

\section{Data}

The dataset comprises 2.4 million patients with 101 million outpatient prescriptions and 58 million diagnoses collected between 2016 and 2022. Women represent 60\% of the population, the median patient age is 47, and patients have a median of 21 prescriptions and 16 diagnoses. Nearly all patients received at least one prescription, and all patients have at least one recorded diagnosis.

Each prescription includes an acute/chronic label and a renewability flag indicating intended treatment duration and dispensing policy. Standard prescriptions may cover three months of treatment, while renewable prescriptions allow repeated dispensing within a one-year period and are primarily used for stable long-term therapy.

Across the dataset, approximately 43 million prescriptions are labeled chronic and 58 million acute, including 11 million chronic-renewable, 32 million chronic non-renewable, 2 million acute-renewable, and 55 million acute non-renewable. Neither administrative field is treated as ground-truth evidence of disease state: acute/chronic labels and renewability reflect prescribing intent and local practice rather than stable clinical phenotypes and should therefore be interpreted as noisy administrative annotations. All prescriptions are mapped to active ingredient level-5 ATC codes to reduce variability due to brand-specific prescribing and ensure consistent aggregation across equivalent formulations.

\section{Methods}

We propose a framework for inferring treated-phenotype onset from longitudinal outpatient prescription data. A treated phenotype denotes a disease state inferred from sustained treatment patterns rather than recorded diagnoses, with onset defined as the transition from sporadic prescribing into sustained outpatient therapy. The framework models prescription timing using renewal-based temporal point processes, detects this transition via likelihood-based change-point analysis between baseline sporadic and sustained renewal regimes, and aggregates drug-level signals into disease-level phenotypes using empirically learned drug--disease associations.

\subsection{Temporal point processes for prescription events}
\label{sec:tpp}

We model prescription timing using renewal processes, which represent events through independent inter-arrival intervals and provide a simple framework for distinguishing sporadic from sustained prescribing behavior. Renewal processes form a restricted class of stochastic temporal point processes (TPPs), in which event timing is modeled through the distribution of inter-arrival intervals rather than fully history-dependent intensity functions~\cite{Aalen2008}. For each patient--drug pair we observe a sequence of prescription times $t_1 < t_2 < \dots < t_{n+1}$ and represent these through inter-arrival times $\tau_i = t_{i+1} - t_i$, for $i = 1, \dots, n$. Prescription events are observed at day-level resolution, and no assumptions are made about medication dispensing or adherence.

We use a homogeneous Poisson process as a null model for sporadic or unstructured prescribing behavior. Under this model the hazard is constant, $h(t)=\lambda$, and inter-arrival times are exponentially distributed:
\begin{equation}
p(\tau \mid \lambda) = \lambda e^{-\lambda \tau}.
\end{equation}
In the context of outpatient prescriptions, this model captures episodic prescribing, acute treatments, or administrative noise and serves as the baseline against which sustained therapy dynamics are evaluated.

To model regular prescription patterns characteristic of sustained outpatient therapy, we use a Weibull renewal process, which has been widely used to model event timing in medical and survival settings~\cite{Pike1966,Foucher2005,Pietzner2013}. Under this model, inter-arrival times follow a two-parameter Weibull distribution:
\begin{equation}
  p(\tau) = \frac{k}{\lambda} \left ( \frac{\tau}{\lambda} \right ) ^{k - 1} e^{-\left ( \tau / \lambda \right )^k},
\end{equation}
where the shape parameter $k$ reflects prescribing regularity ($k \approx 1$ Poisson-like; $k > 1$ regular refill behavior; $k < 1$ bursty patterns) and the scale parameter $\lambda$ determines the characteristic refill interval.

Both parameters $k$ and $\lambda$ admit natural interpretations.
The shape parameter $k$ serves as a \emph{regularity index} of prescribing behavior: $k \approx 1$ indicates Poisson-like, memoryless timing consistent with sporadic prescribing; $k > 1$ reflects increasing temporal regularity characteristic of sustained therapy; and $k < 1$ indicates bursty or clustered events often observed in short-term treatment.

The scale parameter $\lambda$ controls the characteristic time scale between prescriptions. The expected inter-arrival time is $E[\tau] = \lambda \Gamma(1 + 1/k)$. For typical refill regimes ($k \approx 1.5$--$3$), the Gamma term is approximately $0.88$--$0.91$, making $\lambda$ a practical approximation of the refill interval.

Prescription inter-arrival dynamics are strongly influenced by dispensing policy. Renewable prescriptions allow repeated dispensing over extended periods, whereas non-renewable prescriptions typically require more frequent physician authorization.
Pooling these regimes produces distinct time scales and often bimodal inter-arrival distributions.
To account for this, we condition renewal dynamics on the renewability regime, and estimate separate population-level parameters $(k_{d,r}, \lambda_{d,r})$ for each drug $d$ and regime $r \in \{\text{renewable}, \text{non-renewable}\}$:
\begin{equation}
  p_d(\tau \mid r) = \mathrm{Weibull}(\tau \mid k_{d,r}, \lambda_{d,r}).
\end{equation}
These regime-specific estimates are subsequently treated as fixed during patient-level inference of treated-phenotype onsets.

\subsection{Change-point detection for sustained chronic therapy}

We infer the onset of sustained outpatient therapy for a patient--drug pair by modeling prescription timing as a two-regime process with a single change-point. The pre-onset regime corresponds to sporadic prescribing and is modeled by a homogeneous Poisson process, denoted $p_{\text{null}}$, whereas the post-onset regime corresponds to sustained therapy and is modeled using the Weibull renewal process described in Section~\ref{sec:tpp}, denoted $p_{\text{chr}}$.

Given a sequence of inter-arrival times $\tau_1, \dots, \tau_n$ with associated prescribing regimes $r_1, \dots, r_n$, we define the log-likelihood of a candidate change-point location $c$ as
\begin{equation}
\ell(c) =
\sum_{i < c} \log p_{\text{null}}(\tau_i) +
\sum_{c \le i} \log p_{\text{chr}}(\tau_i \mid r_i),
\end{equation}
where $p_{\text{chr}}(\cdot \mid r)$ denotes the regime-specific Weibull renewal model.
The estimated change-point is then obtained as
\begin{equation}
\hat{c} = \arg\max_{c} \ell(c).
\end{equation}
A change-point is accepted only if the log-likelihood exceeds that of the null model $p_{\text{null}}$ applied to the entire sequence by a user-specified threshold $\ell(\hat{c}) - \ell_{\text{null}} > \epsilon$. The parameter $\epsilon$ therefore controls the minimum log-likelihood improvement required to accept a change-point, functioning as a minimum likelihood-ratio evidence threshold rather than a p-value-based statistical test. The calendar date corresponding to $\hat{c}$ is interpreted as the treated-phenotype onset. If no change-point is accepted, no sustained therapy is inferred for that drug.

We restrict inference to a single change-point per patient--drug trajectory, focusing on transitions to sustained outpatient therapy. Clinically, for many chronic indications, the onset of long-term treatment is the primary quantity of interest for cohort construction and leakage mitigation. In contrast, treatment termination is heterogeneous and often not reliably observable in outpatient prescription records: discontinuation may reflect elective stopping, switching therapy, or unobserved events such as death.

Statistically, our inference is based on renewal likelihoods of observed inter-arrival times rather than the full TPP likelihood and therefore does not include survival terms describing the probability of observing no events after the final prescription.
As a result, long post-treatment gaps do not contribute to the objective, and the model captures the transition into sustained therapy but does not attempt to model the termination of therapy.

\subsection{Disease-level treated phenotype construction}

The change-point model produces treated-phenotype onset times for patient--drug pairs. To obtain disease-level treated phenotypes defined over ICD-10 diagnosis categories, we associate drugs with the diseases they treat. In this work, we estimate drug--disease associations empirically through temporal co-occurrence between inferred treated-phenotype onsets and recorded diagnosis events.

For each inferred treated-phenotype onset, we consider a temporal window spanning three months before and 12 months after the onset date, and associate all diagnoses occurring within this window with the corresponding drug.
Treated-phenotype onsets are obtained from the change-point model described above, restricting inference to patients with at least six drug prescriptions and accepting change-points using threshold $\epsilon = 0.05$.

After aggregating associations across patients, we retain drug--ICD pairs with sufficient support ($>25$ associated events).
For each retained pair, we compute a drug-specific alignment rate defined as the number of associated diagnosis events divided by the total number of detected treated-phenotype onsets for that drug.
High alignment rates indicate drugs predominantly prescribed for a specific diagnosis, while low alignment rates indicate non-specific drugs.

For each ICD-10 code, drugs are ranked by alignment rate. We retain drugs with alignment rate $>0.05$, keeping up to the top 30 drugs per ICD code.
Only ICD codes with at least 10 associated drugs are retained.

After defining drug lists for each ICD-10 code, disease-level inference is performed at the patient level. For each patient and ICD code, we consider all drugs in the corresponding drug list and assign the disease treated-phenotype onset as the earliest onset among those drugs.

As a naive baseline, we define treated onset as the first prescription of any disease-associated drug carrying an administrative \emph{chronic} label. This baseline reflects common rule-based phenotyping approaches and is evaluated using the same disease-specific drug sets and patient cohorts as the proposed method. Note that although the change-point detection requires at least six prescriptions to make an inference, the naive baseline requires only a single prescription.

\subsection{Evaluation Protocol}

Patients are randomly split into a training and test set.
The resulting training set comprises 1 million patients with 43 million prescriptions and 25 million diagnoses, while the test set comprises 1.4 million patients with 58 million prescriptions and 34 million diagnoses.
To prevent data leakage, all model components were derived strictly from the training cohort. Specifically, (1) the regime-specific Weibull renewal parameters $(k, \lambda)$ were estimated using training prescriptions, and (2) drug--disease associations were inferred by aligning treated-phenotype onsets with diagnoses solely within the training set. The resulting parameters and dictionary were then frozen and applied to the held-out test cohort to obtain the results reported in Section~\ref{sec:results}.

\section{Results}
\label{sec:results}

\subsection{Drug prescriptions exhibit renewal-like dynamics}

\begin{figure}[t!]
  \centering
  \includegraphics[width=0.9\textwidth]{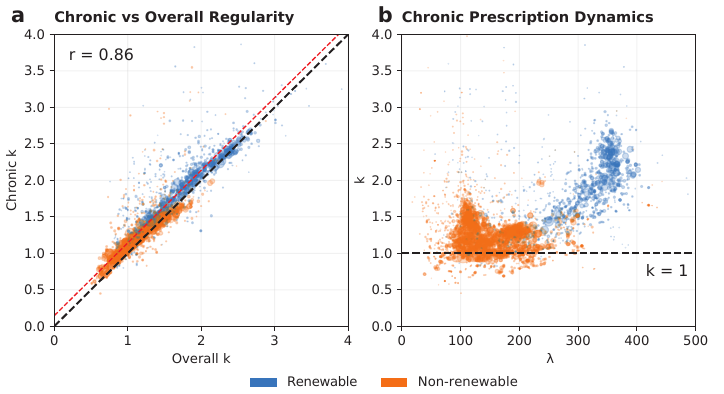}
  \caption{\label{fig:weibull_coefficients}
Weibull parameter estimates across drugs; each point represents one drug. (a) Shape parameter estimates from all prescription intervals versus chronically labeled intervals.  (b) Shape $k$ and scale $\lambda$ stratified by renewable and non-renewable prescriptions; $k>1$ indicates regular (non-Poisson) prescribing. 
Point size reflects the number of intervals.}
\end{figure}

To evaluate robustness to potential noise in administrative chronic/acute labels, we compared Weibull shape parameter $k$ estimates obtained using all prescription intervals versus only intervals labeled as chronic. Fig.~\ref{fig:weibull_coefficients}.a shows strong agreement between the two estimates (Pearson $r=0.86$, slope $0.996$, intercept $0.14$), indicating that renewal regularity structure can be reliably recovered even when chronic/acute labels are noisy, incomplete, or missing.

Although acute prescriptions outnumber chronic prescriptions, they do not typically exhibit consistent renewal structure. While Fig.~\ref{fig:weibull_coefficients}.a suggests that pooling all labels does not introduce systematic bias, acute-labeled prescriptions are more likely to include sporadic prescription events, which can add noise to interval-based parameter estimation. We therefore report results using estimates obtained from chronically labeled prescriptions, acknowledging that although these labels are imperfect ground truth for disease state, they effectively select renewal-type intervals required to estimate $(k, \lambda)$.

Fig.~\ref{fig:weibull_coefficients}.b shows that renewable and non-renewable prescriptions follow markedly different renewal dynamics.
Renewable prescriptions typically exhibit higher scale parameters $\lambda$, often clustering around $\lambda \approx 350$ (roughly one year), indicating longer refill intervals, and higher shape parameter $k$ values, with a substantial cluster above $k=2$, reflecting highly regular long-interval refill schedules once patients enter renewable dispensing regimes.

Consistent with Slovenian national regulatory guidelines, non-renewable prescriptions exhibit shorter refill intervals, with scale parameters typically clustering around $\lambda \approx 100$ days (approximately three months), while still showing predominantly super-Poisson ($k>1$) regularity, albeit with lower $k$ values than renewable prescriptions. This demonstrates that both dispensing regimes exhibit renewal structure but operate on distinct time scales and regularity levels, supporting regime-stratified renewal modeling.

\subsection{Change-point inferred onsets avoid early detections}

\begin{figure}[t!]
  \centering
  \includegraphics[width=\textwidth]{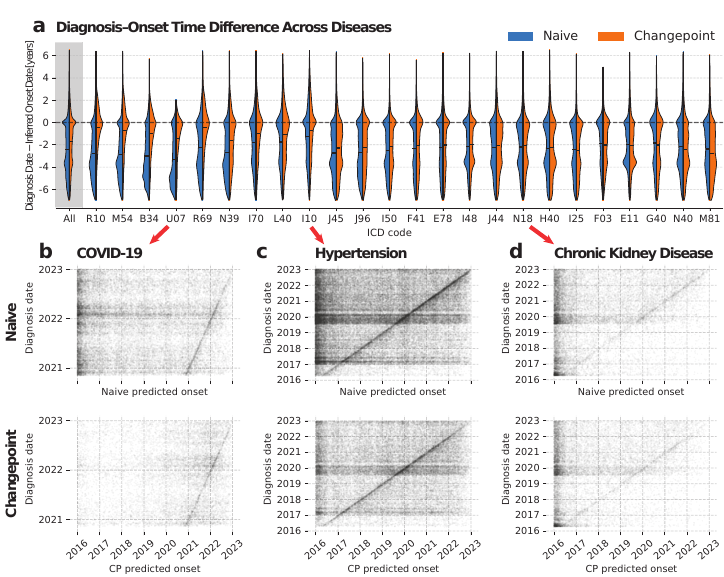}
  \caption{\label{fig:results_time_diff}Distribution of differences between recorded diagnosis date and inferred treated-phenotype onset date for the naive and change-point methods. Violin plots are ordered from left to right by decreasing mean difference. Negative values indicate inferred onset preceding the recorded diagnosis date.}
\end{figure}

Fig.~\ref{fig:results_time_diff}.a shows differences between recorded diagnosis dates and inferred treated-phenotype onset dates. Across diseases, both methods often infer onset preceding recorded diagnoses, consistent with left-censored EHR data; however, excessively early detections suggest over-triggering. Compared to the naive baseline, the change-point method produces tighter distributions centered near zero, indicating improved temporal alignment. Example panels in Figs.~\ref{fig:results_time_diff}.b--d show a diagonal band indicating agreement between inferred onset and diagnosis timing. In contrast, early-onset bands near the start of data collection suggest ongoing treatment at EHR adoption and are more pronounced for the naive method.

We can further verify this using ICD code U07 (COVID-19), which provides a natural temporal sanity check, as the disease did not exist prior to 2019. Although COVID-19 is not a chronic disease, administrative prescribing labels reflect dispensing policy and workflow rather than disease chronicity and may therefore appear in this analysis. The naive method predicts a substantial number of COVID-19 onsets as early as 2016, whereas the change-point method largely avoids such implausible early detections. This indicates that the naive method over-triggers onset events relative to the change-point approach.

\subsection{Change-point detection is limited by prescription density}

\begin{figure}[t!]
  \centering
  \includegraphics[width=\textwidth]{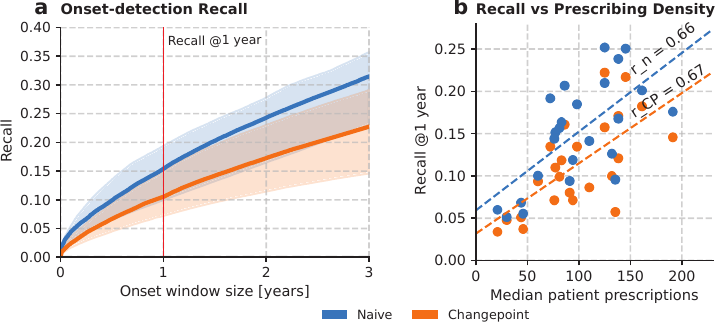}
  \caption{\label{fig:recall}Disease-level recall of naive and renewal-based onset detection.  (a) Recall as a function of the symmetric temporal tolerance window around the diagnosis date. Curves show the median recall across ICD codes and shaded regions show the interquartile range (IQR). (b) Relationship between disease-level recall in the $\pm 365$ day window around the true diagnosis date and prescription density, measured as the median number of prescriptions per diagnosed patient. Each point represents one ICD code.}
\end{figure}

Fig.~\ref{fig:recall} summarizes disease-level detection performance of the naive and renewal-based onset detection methods. To evaluate time-aware detection accuracy, we define a temporal tolerance window $\Delta$ and count a detection as correct if the inferred onset occurs within $\pm \Delta$ days of the recorded diagnosis date. Fig.~\ref{fig:recall}.a shows recall as a function of $\Delta$. Across diseases, the renewal-based method consistently shows lower recall than the naive baseline, reflecting its more conservative triggering behavior. Recall increases for both methods as $\Delta$ increases, approaching their respective maximum detection coverage for larger tolerance windows.

Recall varies across ICD codes and is associated with prescription density. Fig.~\ref{fig:recall}.b shows disease-level recall at $\Delta = 365$ days as a function of the median number of prescriptions per diagnosed patient.
Diseases with denser longitudinal prescribing patterns exhibit substantially higher recall for both methods (TPP: $r=0.67$, naive: $r=0.66$).
This indicates that detection performance is primarily constrained by the availability of longitudinal prescribing signal rather than model-specific limitations. Because the naive approach requires only a single prescription to trigger an onset event, it naturally achieves higher recall than the change-point detection method, which requires sustained evidence of longitudinal treatment. Consistent with this, the naive approach generates substantially more onset detections overall (3.1 million vs. 1.5 million for the change-point method).

Overall, the change-point method is intentionally more conservative than the naive baseline because it requires sustained longitudinal prescribing evidence before triggering onset. As a result, diseases with sparse or episodic prescribing are harder to detect from treatment data, while dense chronic trajectories yield more reliable detection. Because diagnosis coding is incomplete and not missing at random (e.g., external care or pre-EHR diagnoses), recall is reported relative to recorded diagnoses and should be interpreted as a lower bound on detection coverage rather than an unbiased estimate of true prevalence.

\section{Conclusion}

We show that longitudinal prescribing data contain recoverable signal about treated-phenotype onset and that renewal-based change-point detection provides temporally plausible onset estimates under strong left censoring. Compared to naive rule-based triggering, the proposed method reduces implausible early detections by requiring sustained treatment evidence, improving cohort construction despite incomplete diagnosis coding. Detection performance depends strongly on prescribing density: diseases with structured chronic therapy yield reliable signals, whereas sparse or episodic prescribing remains challenging. These findings highlight the importance of modeling treatment dynamics in EHR phenotyping and motivate future extensions incorporating treatment switching and external biomedical knowledge.

\begin{credits}
\subsubsection{\ackname}
Our research was supported by the grant from the Slovenian Research and Innovation Agency (grant no. P2-0209). We would also like to thank the Slovenian National Institute of Public Health for their constructive cooperation.

\end{credits}

\bibliographystyle{splncs04}
\bibliography{main}
\end{document}